\documentclass[conference]{IEEEtran}
\IEEEoverridecommandlockouts
\usepackage{cite}
\usepackage{amsmath,amssymb,amsfonts}
\usepackage{algorithmic}
\usepackage{graphicx}
\usepackage{textcomp}
\usepackage{xcolor}
\usepackage[font=small,labelsep=period]{caption}
\usepackage{multirow}
\DeclareCaptionFont{eightpt}{\fontsize{8}{9.6}\selectfont}

\def\BibTeX{{\rm B\kern-.05em{\sc i\kern-.025em b}\kern-.08em
    T\kern-.1667em\lower.7ex\hbox{E}\kern-.125emX}}
\begin{document}

\title{SEDR-Seq2P: A Lightweight Dilated Residual Sequence-to-Point Network for Multi-Task Industrial NILM}

\author{
\IEEEauthorblockN{Hatem Haddad}
\IEEEauthorblockA{\textit{Wattnow}\\
Tunis, Tunisia \\
hatem.haddad@wattnow.io}
\and
\IEEEauthorblockN{Feres Jerbi}
\IEEEauthorblockA{\textit{Wattnow}\\
Tunis, Tunisia \\
feres.jerbi@wattnow.io}
\and
\IEEEauthorblockN{Issam Smaali}
\IEEEauthorblockA{\textit{Wattnow}\\
Tunis, Tunisia \\
issam@wattnow.io}
}

\IEEEoverridecommandlockouts
\IEEEpubid{\makebox[\columnwidth]{979-8-3195-1058-7/26/\$31.00~\copyright2026 IEEE \hfill} 
\hspace{\columnsep}\makebox[\columnwidth]{ }}
\maketitle

\begin{abstract}
Industrial NILM remains challenging because measurement noise and widespread concurrent machine operation reduce the generalization of models tuned on residential data. This work adopts a one-to-many, multi-task disaggregation setting, in which a single network estimates multiple industrial machine loads from aggregate power. Under a unified evaluation protocol on IMDELD, we benchmark Seq2Seq, Seq2SubSeq, Seq2Point, GRU, and WaveNet using energy-estimation metrics and the accuracy–delay criterion. While Seq2Point offers a stronger accuracy–delay balance than Seq2Seq/Seq2SubSeq, GRU and WaveNet achieve higher accuracy at markedly higher computational cost. To close this gap, we propose SEDR-Seq2P, a lightweight Seq2Point extension with dilated residual blocks and squeeze-and-excitation attention. Relative to the Seq2Point baseline, SEDR-Seq2P reduces MAE by approximately 7\%, improves the coefficient of determination by approximately 1\%, and increases the match rate by approximately 0.8\%. In addition, compared to WaveNet, SEDR-Seq2P reduces inference latency by approximately 58\%, yielding a favorable accuracy–delay trade-off for scalable industrial deployment.
\end{abstract}

\section{Introduction}
Appliance Load Monitoring (ALM) enables fine-grained analysis of energy consumption in industrial equipment, supporting operational efficiency, energy optimization, and predictive maintenance within Industry 4.0 frameworks. Recent advances in sensing technologies and AI-based analytics have improved the accuracy and scalability of industrial ALM by enabling the identification of high-consumption machines and informed load management strategies. Appliance-level monitoring can be achieved through Intrusive Load Monitoring (ILM) or Non-Intrusive Load Monitoring (NILM) \cite{b8}; however, ILM relies on dedicated per-device sensors, leading to high installation and maintenance costs and limiting its suitability for large-scale industrial deployment. NILM refers to a class of signal processing and machine learning techniques that estimate appliance-level electricity consumption from aggregate power measurements acquired at a limited number of points in a building’s electrical distribution system, typically at the mains. By relying solely on aggregate measurements, NILM enables whole-building load disaggregation without the need for dedicated sensors at individual appliances.
Although NILM research dates back to 1985 \cite{b9}, most disaggregation methods have been evaluated predominantly on residential benchmarks \cite{b10}, largely due to the scarcity of publicly available industrial datasets. However, prior studies \cite{b10} show that strong residential performance does not necessarily generalize to industrial environments, which differ fundamentally in their operational characteristics, including shift-based production schedules, a wide array
of complex machinery operating under demanding and noisier
conditions, and high levels of concurrent machine operation that violate common residential NILM assumptions \cite{Yaniv}. Consequently,  the “one-at-a-time” appliance usage
model often applied in domestic NILM is generally inapplicable in industrial contexts, where simultaneous operation of multiple devices is
the norm.
Energy disaggregation in industrial environments is inherently unidentifiable and can be formulated as a single-channel blind source separation problem \cite{hart}. In this setting, multiple loads must be inferred from a single aggregate power signal under noise, overlapping consumption patterns, and frequent concurrent operation. As a result, purely deterministic NILM models, such as combinatorial optimization, deterministic HMMs, linear regression, non-negative matrix factorization, and rule-based approaches, produce a single fixed solution. Consequently, they fail to represent the intrinsic ambiguity of industrial disaggregation tasks \cite{Zhang}. To address this, sequence-to-sequence (Seq2Seq) learning was proposed to model single-channel load disaggregation by mapping aggregate power sequences to appliance-level sequences \cite{Kelly}. However, long input-output sequences incur high computational costs and can hinder convergence. Sequence-to-Point (Seq2Point) learning \cite{Zhang} mitigates these limitations by predicting only the midpoint of an output window, reducing training complexity while preserving temporal context. Sequence-to-Subsequence (Seq2SubSeq) \cite{subseq} represents an intermediate learning paradigm between Seq2Seq and Seq2Point, designed to balance temporal context and prediction efficiency in NILM. It maps a window of aggregate mains power to a short, typically centered, subsequence of appliance-level power. Compared to Seq2Seq, Seq2SubSeq reduces redundant predictions and edge-smoothing effects, while preserving more local temporal structure than Seq2Point, which predicts only a single time step.

This work adopts a one-to-many industrial NILM setting rather than training a separate model per device. This multi-task formulation better matches concurrent industrial operation and reduces deployment overhead. The key contributions of this paper are four-fold:
\begin{itemize}
\item Multi-task industrial NILM: We adopt a one-to-many setting in which a single shared model jointly disaggregates all machines from aggregate mains power, matching concurrent operation in industrial sites.
\item Unified benchmarking and evaluation: We benchmark Seq2Seq, Seq2SubSeq, Seq2Point, GRU, and WaveNet on IMDELD dataset under identical preprocessing and training conditions, reporting energy-estimation (EE) metrics, namely mean absolute error (MAE), normalized disaggregation error (NDE), signal aggregate error (SAE), coefficient of determination ($R^2$), and event-level temporal alignment via match rate (MR).
\item Accuracy-Delay assessment: To jointly assess predictive quality and computational efficiency, we adopt an  Accuracy–Delay criterion (AccD) \cite{ARR3} derived from the adjusted ratio of ratios (ARR) framework. In this work, MR is used as the accuracy term and measured test-time inference duration is used as the delay term. %We complement EE results with adjusted ratio of ratios (ARR) \cite{ARR2}, combining match rate with measured inference time to quantify the accuracy–delay trade-off for industrial deployment.
\item Proposed lightweight dilated residual Seq2Point architecture:
We introduce SEDR-Seq2P, a structurally enhanced Seq2Point model that integrates dilated residual blocks and squeeze-and-excitation channel attention within a multi-task one-to-many framework, explicitly designed to improve temporal modeling capacity without sacrificing inference efficiency.
%\item Accuracy--complexity trade-off: We show that SEDR-Seq2P improves over sequence models while remaining substantially more efficient than heavier recurrent and WaveNet baselines.
\end{itemize}
EE metrics assess disaggregation accuracy independently of computational cost, whereas the AccD metric captures the trade-off between accuracy and computation time. AccD therefore complements rather than replaces EE metrics, as practical industrial NILM deployment requires balancing predictive performance with computational efficiency. In this work, training and inference times are jointly evaluated with accuracy using the AccD parameter to assess overall model efficiency. First, we report the training time per epoch and the total inference time on the test set for all models. We then combine predictive accuracy and computational performance using the AccD metric based on MR.
\section{NILM Benchmark Models}

\begin{figure*}[htbp]
\centering
\includegraphics[width=0.90\textwidth]{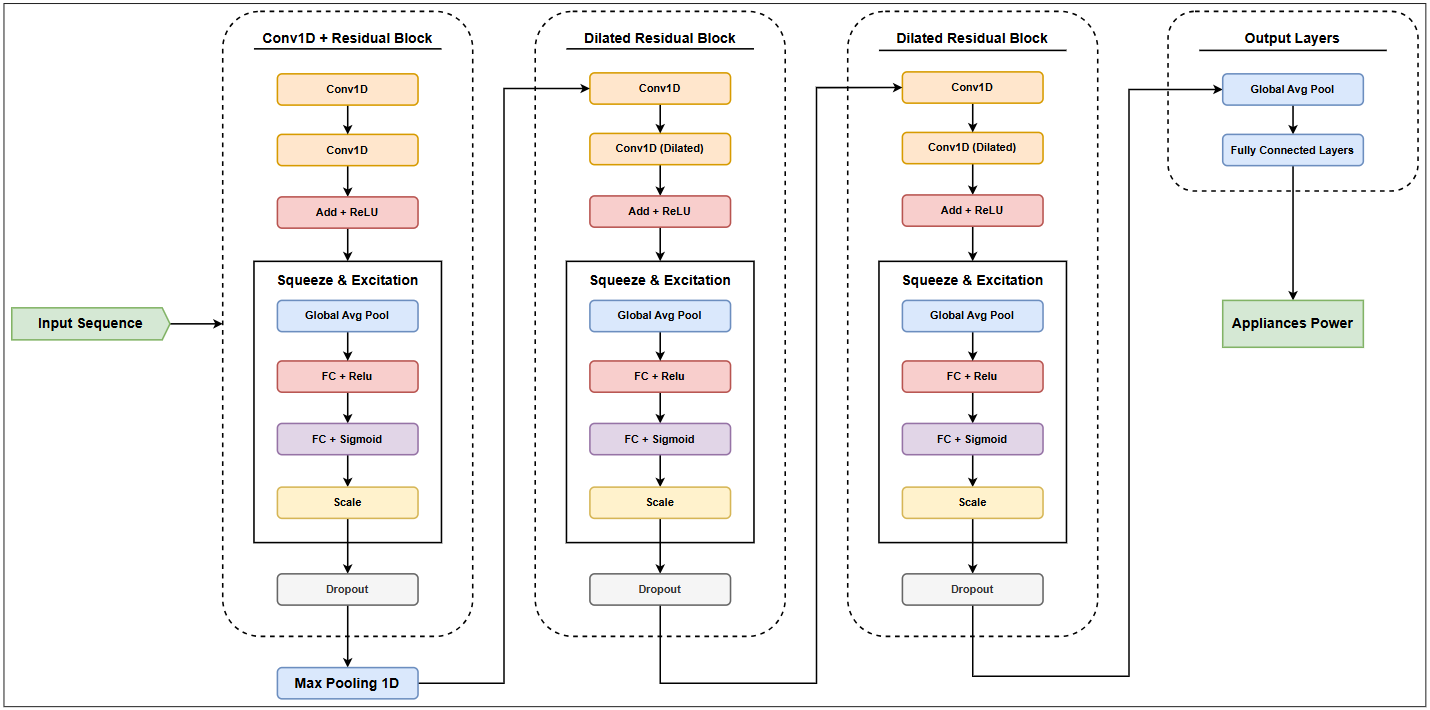}
\caption{Overall architecture of the proposed model. FC denotes fully connected layers.}
\label{structure}
\end{figure*}

In this work, the following learning-based NILM models are evaluated:
\begin{itemize}
\item Seq2Point \cite{Zhang}: maps a fixed-length window of aggregate mains power to a single appliance-power estimate at the window center, which limits redundant overlap predictions and typically reduces inference cost.
\item Seq2Seq \cite{Kelly}: learns an end-to-end mapping from an aggregate input window to a full appliance-level output window, preserving temporal structure but increasing computation due to dense, overlapping sequence predictions.
\item Seq2SubSeq \cite{subseq}: predicts a short centered segment of the appliance signal from each input window, providing an intermediate trade-off between temporal detail and prediction efficiency.
\item Gated recurrent unit networks (GRU) \cite{ARR}: use recurrent gating to model long-range temporal dependencies in appliance operation, which can improve tracking of prolonged and variable industrial cycles but often increases runtime.
\item WaveNet-style temporal convolutional networks \cite{Jiang,Vavouris}: employ stacks of dilated convolutions (often with residual connections) to expand the receptive field efficiently and capture both sharp switching events and sustained operating plateaus.
\end{itemize}
In a second step, motivated by the comparative performance of the evaluated models (Table~\ref{tab:imdedl_comparison}), we introduce a Seq2Point architecture incorporating residual blocks \cite{residual} and dilated convolutional layers \cite{dilated} to better capture long-range temporal dependencies. In addition, Squeeze-and-Excitation channel attention \cite{SE} is integrated to strengthen the representation of the most informative feature channels, as illustrated in Fig.~\ref{structure}. The resulting architecture is referred to as SEDR-Seq2P.

\section{Experimental Setup}
\subsection{IMDELD Dataset}

The IMDELD (Industrial Machines Dataset for Electrical Load Disaggregation) \cite{imdeld} is a publicly available industrial dataset collected using eleven GreenAnt energy meters (one site meter and ten meters for ground-truth monitoring). Each meter samples internally at 8 kS/s and provides a 1-Hz downsampled time series, which we use without further resampling. The dataset covers eight machines: double-pole contactors (DPCI, DPCII), exhaust fans (EFI, EFII), milling machines (MMI, MMII), and pelletizers (PI, PII). The dataset provides RMS voltage, current, active, reactive, and apparent power, as well as active energy measurements, with all inputs normalized using z-score statistics \cite{zscore} computed on the training set. We conducted a feature-correlation analysis using Pearson, Kendall, and Spearman coefficients \cite{Correlation}, which showed that RMS voltage, RMS current, reactive power, and apparent power are highly correlated with active power; therefore, only active power is used as the input feature. 

The dataset covers 111 days; however, the final 3 days were excluded due to missing appliance-level measurements. From the remaining data, 7 days were used for validation and 1 day for testing, resulting in 691,072 test samples per appliance, while the remaining data were used for training.

\subsection{Hyperparameters}

All deep learning models are trained using the Adam optimizer and MAE loss function. Hyperparameters are optimized via random search with 20 trials per model sampled from a predefined grid. Training is performed for up to 30 epochs with a batch size of 128, using early stopping based on validation performance with a patience of 5 epochs, and a ReduceLROnPlateau scheduler to adapt the learning rate during training.
\begin{table}[htbp]
\caption{Selected Hyperparameters for Sequence Models on IMDELD}
\centering
\small
\begin{tabular}{|c|c|c|c|}
\hline
\textbf{Parameter} 
& \textbf{Seq2Seq} 
& \textbf{Seq2SubSeq} 
& \textbf{Seq2Point} \\
\hline
\textbf{Filters 1}        & 64  & 32  & 32  \\
\hline
\textbf{Filters 2}        & 128 & 64  & 128 \\
\hline
\textbf{Filters 3}        & 128 & 256 & 256 \\
\hline
\textbf{Kernel\_Size}    & 7   & 7   & 3   \\
\hline
\textbf{Dense 1}          & 256 & 256 & 256 \\
\hline
\textbf{Dense 2}          & 64  & 128 & 64  \\
\hline
\textbf{Dropout}         & 0.3 & 0.3 & 0.3 \\
\hline
%\textbf{Learning\_Rate}  & 0.001 & 0.0005 & 0.001 \\
%\hline
\end{tabular}
\label{tab:imdel_best_params_seq2}
\end{table}
As the proposed multi-task framework requires a common window size across appliances, we adopt a unified input window length of 155 samples for all sequence models. This length is set to the mean of the appliance-specific optimal window sizes reported in \cite{b12}, obtained using LightGBM in a single-task setting. For sequence-to-subsequence architectures, the output subsequence length is fixed at 75 samples. The sequence models hyperparameters selected through random search are summarized in Table~\ref{tab:imdel_best_params_seq2}.

Although the GRU and WaveNet models are tuned using the same training setup, they are omitted from the table due to architectural differences. The best-performing GRU configuration uses 32 recurrent units followed by two fully connected layers with 128 and 64 units and a dropout rate of 0.2. The best-performing WaveNet configuration uses three convolutional layers with 64, 128, and 256 filters (kernel size 5), two residual blocks with four dilated convolutional layers per block, a dropout rate of 0.3, and a learning rate of 0.001. 

The SEDR-Seq2P model uses an initial Conv1D layer with 32 filters, followed by three residual blocks with 32, 128, and 256 filters, respectively; the last two blocks use dilations of 2 and 4. We use a kernel size of 3, a Squeeze-and-Excitation reduction ratio of 0.25, a dropout rate of 0.3, and a dense head with 256 and 64 units, trained with a learning rate of $10^{-3}$. 

All experiments are conducted on an NVIDIA GeForce RTX 3070 Laptop GPU with 8 GB of memory.

\subsection{Performance Evaluation}
For performance assessment, we employ five complementary EE metrics \cite{Mayhorn}: MAE, which measures the average absolute deviation between predicted and ground-truth power; NDE to quantify per-appliance estimation accuracy; SAE to assess consistency of total estimated energy; $R^2$ to capture trend and magnitude agreement; and MR to evaluate temporal alignment between predicted and ground-truth activations. We emphasize MR because it captures activation timing and overlap, which are critical in industrial NILM under frequent concurrent operation and complex temporal dynamics.

In addition, an accuracy--delay (AccD) efficiency score \cite{ARR3} is employed to jointly evaluate disaggregation accuracy and computational efficiency in multi-appliance settings. In our experiments, the Match Rate (MR) is used as the success-rate term because it directly quantifies correct detection of appliance operating intervals and is bounded in $[0,1]$, facilitating consistent aggregation across appliances. Accordingly, we define the following $\mathrm{AccD}_{\mathrm{MR}}$ for multi-appliance evaluation:
\begin{equation}
\mathrm{AccD}_{\mathrm{MR}} =
\frac{1}{T}
\cdot
\frac{1}{|\mathcal{A}|}
\sum_{i \in \mathcal{A}} \mathrm{MR}_i ,
\end{equation}
where $\mathcal{A}$ denotes the set of appliances, $\mathrm{MR}_i$ is the Match Rate of appliance $i$, and $T$ is the total inference time measured over the test dataset under identical hardware conditions. This formulation provides a relative criterion for assessing deployability by jointly accounting for temporal accuracy and inference latency.

\section{Experimental Results}
\begin{table*}[htbp]
\caption{Comparison of energy-estimation performance across models}
\centering
\scriptsize
\begin{tabular}{|l|l|c|c|c|c|c|c|c|c|c|}
\hline
\textbf{Model} & \textbf{Metric} & \textbf{MM I} & \textbf{MM II} & \textbf{Pel I} & \textbf{Pel II} & \textbf{DPC I} & \textbf{DPC II} & \textbf{EF I} & \textbf{EF II} & \textbf{Average} \\
\hline

\multirow{5}{*}{\textbf{Seq2Seq}}
 & \textbf{MAE} & 5696.0610 & 5377.2207 & 5432.5160 & 6828.9536 & 128.4504 & 130.5084 & 153.6593 & 240.9586 & 2998.5410 \\
 & \textbf{NDE} & 0.1100 & 0.1145 & 0.0414 & 0.0516 & 0.0689 & 0.0654 & 0.0417 & 0.0282 & 0.0652 \\
 & \textbf{SAE} & 0.0524 & 0.0356 & 0.0227 & 0.0213 & 0.0816 & 0.0802 & \textbf{0.0061} & 0.0017 & 0.0377 \\
 & \textbf{$R^2$} & 0.7685 & 0.7525 & 0.8489 & 0.8260 & 0.7625 & 0.7754 & 0.8350 & 0.8838 & 0.8066 \\
 & \textbf{MR} & 0.7937 & 0.7920 & 0.9141 & 0.8846 & 0.8417 & 0.8453 & 0.9405 & 0.9450 & 0.8696 \\
\hline

\multirow{5}{*}{\textbf{Seq2Point}}
 & \textbf{MAE} & 5293.8774 & 4959.2390 & 5266.5100 & 6352.7007 & 130.5998 & 131.2622 & 150.1366 & 225.8590 & 2813.7730 \\
 & \textbf{NDE} & 0.0886 & 0.0940 & 0.0363 & 0.0493 & 0.0725 & 0.0666 & 0.0409 & 0.0260 & 0.0593 \\
 & \textbf{SAE} & 0.0476 & 0.0346 & 0.0415 & 0.0293 & \textbf{0.0511} & 0.0543 & 0.0169 & 0.0065 & 0.0352 \\
 & \textbf{$R^2$} & 0.8135 & 0.7968 & 0.8675 & 0.8339 & 0.7500 & 0.7714 & 0.8381 & 0.8930 & 0.8205 \\
 & \textbf{MR} & 0.8073 & 0.8065 & 0.9174 & 0.8918 & 0.8371 & 0.8427 & 0.9421 & 0.9485 & 0.8742 \\
\hline

\multirow{5}{*}{\textbf{Seq2SubSeq}}
 & \textbf{MAE} & 5691.1016 & 5379.8687 & 5482.6025 & 6761.1646 & 127.1573 & 128.3626 & 156.1064 & 235.8518 & 2995.2769 \\
 & \textbf{NDE} & 0.1158 & 0.1197 & 0.0439 & 0.0531 & 0.0715 & 0.0681 & 0.0430 & 0.0280 & 0.0679 \\
 & \textbf{SAE} & 0.0316 & 0.0558 & 0.0351 & 0.0178 & 0.0724 & 0.0713 & 0.0073 & \textbf{0.0013} & 0.0366 \\
 & \textbf{$R^2$} & 0.7564 & 0.7412 & 0.8399 & 0.8210 & 0.7537 & 0.7661 & 0.8297 & 0.8850 & 0.7991 \\
 & \textbf{MR} & 0.7959 & 0.7937 & 0.9138 & 0.8859 & 0.8425 & 0.8470 & 0.9396 & 0.9462 & 0.8706 \\
\hline

\multirow{5}{*}{\textbf{GRU}}
 & \textbf{MAE} & 4748.5010 & 4024.6465 & \textbf{5177.2593} & \textbf{5755.1904} & 121.1604 & 123.5602 & \textbf{142.9064} & 232.3958 & 2699.2683 \\
 & \textbf{NDE} & 0.0727 & \textbf{0.0680} & \textbf{0.0355} & \textbf{0.0389} & \textbf{0.0559} & \textbf{0.0519} & \textbf{0.0362} & 0.0279 & 0.0489 \\
 & \textbf{SAE} & 0.0720 & \textbf{0.0240} & 0.0436 & 0.0217 & 0.0745 & 0.0765 & \textbf{0.0061} & 0.0019 & \textbf{0.0294} \\
 & \textbf{$R^2$} & 0.8469 & \textbf{0.8531} & \textbf{0.8703} & \textbf{0.8688} & 0.8073 & \textbf{0.8217} & 0.8567 & 0.8852 & 0.8494 \\
 & \textbf{MR} & 0.8234 & 0.8393 & \textbf{0.9188} & \textbf{0.9019} & 0.8495 & 0.8527 & \textbf{0.9446} & 0.9469 & 0.8821 \\
\hline

\multirow{5}{*}{\textbf{WaveNet}}
 & \textbf{MAE} & 4710.6330 & 4440.1733 & 5512.1880 & 6272.8657 & \textbf{114.9032} & \textbf{110.6361} & 155.5892 & 277.1583 & \textbf{2540.7024} \\
 & \textbf{NDE} & \textbf{0.0685} & 0.0699 & 0.0376 & 0.0437 & 0.0589 & 0.0538 & \textbf{0.0362} & \textbf{0.0228} & \textbf{0.0484} \\
 & \textbf{SAE} & \textbf{0.0230} & 0.0371 & \textbf{0.0183} & 0.0229 & 0.0583 & \textbf{0.0491} & 0.0068 & 0.0195 & 0.0400 \\
 & \textbf{$R^2$} & \textbf{0.8558} & 0.8488 & 0.8628 & 0.8526 & 0.7969 & 0.8154 & \textbf{0.8568} & \textbf{0.9062} & \textbf{0.8513} \\
 & \textbf{MR} & 0.8287 & 0.8252 & 0.9127 & 0.8934 & \textbf{0.8557} & \textbf{0.8654} & 0.9394 & 0.9364 & \textbf{0.8846} \\
\hline
\multirow{5}{*}{\textbf{SEDR-Seq2P}}
 & \textbf{MAE} & \textbf{4566.5146} & \textbf{3907.8904} & 5466.5024 & 6290.8003 & 134.5565 & 133.3088 & 146.7158 & \textbf{215.8484} & 2607.7670 \\
 & \textbf{NDE} & 0.0737 & 0.0697 & 0.0441 & 0.0510 & 0.0718 & 0.0655 & 0.0395 & 0.0249 & 0.0550 \\
 & \textbf{SAE} & 0.0799 & 0.0258 & 0.0379 & 0.0390 & 0.0755 & 0.0719 & 0.0164 & 0.0112 & 0.0447 \\
 & \textbf{$R^2$} & 0.8450 & 0.8493 & 0.8391 & 0.8282 & 0.7525 & 0.7750 & 0.8435 & 0.8974 & 0.8288 \\
 & \textbf{MR} & \textbf{0.8290} & \textbf{0.8437} & 0.9142 & 0.8923 & 0.8344 & 0.8417 & 0.9434 & \textbf{0.9509} & 0.8812 \\
\hline
\end{tabular}
\label{tab:imdedl_comparison}
\end{table*}

\subsection{EE Performances}
Table~\ref{tab:imdedl_comparison} indicates that disaggregation performance is strongly influenced by architectural design, particularly the ability to capture temporal dependencies at multiple scales. Seq2Point consistently outperforms Seq2Seq and Seq2SubSeq, achieving a lower average MAE (2813.77 vs.\ 2998.54 and 2995.28) and a higher average $R^2$ (0.8205 vs.\ 0.8066 and 0.7991). For high-power machines, Seq2Point reduces NDE by approximately 18\%--19\% for milling machines and 4\%--12\% for pelletizers relative to Seq2Seq, and by 21\%--23\% and 7\%--17\%, respectively, relative to Seq2SubSeq. These results are consistent with the Seq2Point formulation, which concentrates representational capacity on a single central prediction rather than distributing it across an output sequence. By avoiding overlapping output windows, Seq2Point reduces prediction redundancy and helps reduce boundary smoothing effects introduced by overlapping window predictions, leading to improved preservation of activation transitions and more accurate amplitude estimation. This structural characteristic appears particularly beneficial for high-power equipment exhibiting sustained operating states and abrupt switching events.

GRU provides consistent performance improvements over purely convolutional sequence models, reducing the average MAE to 2699.27 and the average NDE to 0.0489, while increasing the average $R^2$ to 0.8494. The gains are more pronounced for milling machines with long and variable operating cycles, suggesting that modeling long-term temporal dependencies is beneficial for industrial equipment disaggregation. WaveNet attains the highest overall accuracy among the evaluated models, achieving the lowest average MAE (2540.70) and the highest average $R^2$ (0.8513). Its use of dilated causal convolutions and residual connections expands the receptive field efficiently, which may facilitate the modeling of both switching events and sustained power levels. The resulting $R^2$ values for exhaust fans further support the relevance of multi-scale temporal feature extraction for appliances exhibiting both steady-state and transient behavior.

MR analysis complements these findings by assessing event-level temporal alignment. Convolutional sequence models exhibit comparatively lower detection performance, whereas GRU and WaveNet-based architectures achieve improved event alignment. This trend suggests that capturing long-range temporal dependencies while preserving sensitivity to abrupt transitions contributes to more accurate event detection in industrial NILM.

When evaluated using $\mathrm{AccD}_{\mathrm{MR}}$ (Table~\ref{tab:accd_mr_comparison}), the models exhibit distinct trade-offs between temporal accuracy and inference latency. Although GRU and WaveNet achieve higher average MR, their longer inference times result in lower $\mathrm{AccD}_{\mathrm{MR}}$  values. In contrast, Seq2Point combines competitive MR performance with the lowest inference time, yielding the highest $\mathrm{AccD}_{\mathrm{MR}}$  among the evaluated models. These results indicate that improvements in temporal accuracy alone do not necessarily translate into higher overall efficiency when computational cost is considered. Motivated by this observation, we propose SEDR-Seq2P, a hybrid extension of Seq2Point designed to improve predictive accuracy while preserving a lightweight inference profile. 
%The proposed architecture enhances the baseline Seq2Point model by integrating residual blocks and dilated convolutional layers to capture extended temporal context, together with Squeeze-and-Excitation channel attention to emphasize informative feature representations.
\begin{table}[htbp]
\caption{$\mathrm{AccD}_{\mathrm{MR}}$ performances. Training time is reported per epoch, and inference time is reported as total test-set inference time in seconds.}
\centering
\begin{tabular}{|c|c|c|c|c|}
\hline
\textbf{Model} & \textbf{Avg. MR}  & \textbf{Train } & \textbf{Inference} & \textbf{AccD$_{\mathrm{MR}}$} \\
\hline
\textbf{Seq2Seq}   & 0.8696  & \textbf{33.18} & 7.88  & 0.110 \\
\hline
\textbf{Seq2Point}  & 0.8742 & 52.86 & \textbf{6.47}  & \textbf{0.135} \\
\hline
\textbf{Seq2SubSeq}  & 0.8706 & 47.41 & 7.95  & 0.109 \\
\hline
\textbf{GRU}   & 0.8821 & 95.37 & 25.09 & 0.035 \\
\hline
\textbf{WaveNet}  & \textbf{0.8846}  & 270.23 & 23.56 & 0.037 \\
\hline
\textbf{SEDR-Seq2P} & 0.8812 & 45.72 & 9.88 & 0.089 \\
\hline
\end{tabular}
\label{tab:accd_mr_comparison}
\end{table}
\subsection{SEDR-Seq2P Performances}

SEDR-Seq2P improves upon the Seq2Point baseline across multiple evaluation criteria, reducing NDE by approximately 7\%, increasing $R^2$ by about 1\%, and improving MR by 0.8\% on average. Although these gains are moderate, they are consistent across appliances, indicating systematic performance improvements. In terms of efficiency, SEDR-Seq2P remains significantly faster than GRU and WaveNet, but does not achieve the highest AccD$_{\mathrm{MR}}$, which is retained by Seq2Point. Compared with heavier baselines, however, it provides a more favorable accuracy--efficiency trade-off, reducing inference latency by approximately 50\% relative to WaveNet while maintaining competitive accuracy.

The observed improvements are consistent with the architectural extensions introduced in SEDR-Seq2P, including dilated convolutions and residual connections, which increase the effective temporal receptive field and support deeper feature extraction without substantially increasing inference complexity. From a NILM perspective, the enlarged receptive field may facilitate improved modeling of both short-duration switching events and longer steady-state operating periods, contributing to enhanced event detection while preserving steady-load energy estimation accuracy.

\subsection{Statistical significance analysis}
Statistical significance was evaluated using the Friedman test
followed by Wilcoxon signed-rank tests with Holm correction,
as recommended for multiple learning algorithms comparisons \cite{statisticaltests}.
The Friedman test indicates significant differences between
models for MR ($\chi^2=13.07$, $p=0.0227$) and MAE
($\chi^2=12.57$, $p=0.0277$).
However, pairwise comparisons reveal that these accuracy
differences are not statistically significant after multiple
comparison correction, suggesting that most models achieve
comparable predictive performance.

In contrast, AccD$_{\mathrm{MR}}$ shows highly significant differences across models ($\chi^2 = 40.0$, $p < 0.001$). Pairwise Wilcoxon signed-rank tests with Holm correction indicate that SEDR-Seq2P achieves significantly better efficiency scores than the heavier GRU and WaveNet baselines (Holm-corrected $p = 0.039$), while the highest AccD$_{\mathrm{MR}}$ is still obtained by the original Seq2Point model. These results suggest that SEDR-Seq2P provides a competitive accuracy--efficiency trade-off for industrial NILM deployment.

\section{State of the Art}

Early NILM research formulated load disaggregation as a non-intrusive inference problem based on aggregate power measurements~\cite{hart}. With the adoption of deep learning, Seq2Seq learning was introduced to map aggregate input windows to appliance-level output sequences~\cite{Kelly}. While effective, long input-output horizons increase computational costs and may complicate optimization due to redundant overlapping predictions. Seq2Point learning~\cite{Zhang} addressed this limitation by predicting only the midpoint sample of each window, thereby reducing redundancy while preserving sufficient temporal context. Seq2SubSeq learning~\cite{subseq} further explored an intermediate strategy by predicting short centered output segments to balance contextual modeling and computational efficiency.

In industrial settings, the IMDELD dataset was first investigated in~\cite{b13}, where FHMM and WaveNet-based approaches were evaluated for machinery disaggregation. Subsequent work highlighted that methods performing well on residential datasets often exhibit limited generalization in industrial environments due to sustained operating cycles, high concurrency, and stronger measurement noise~\cite{b10,Yaniv}. These characteristics challenge architectures that assume sparse or sequential appliance activations.

To address long-range dependencies, recurrent neural networks such as GRU have been adopted for NILM, improving modeling of prolonged and variable load patterns at the expense of increased runtime~\cite{ARR}. Temporal convolutional networks, including WaveNet-style architectures~\cite{Jiang,Vavouris}, employ stacks of dilated convolutions and residual connections to efficiently enlarge the receptive field and capture both sharp switching events and steady-state plateaus. Although such models achieve strong predictive accuracy, their computational cost and architectural depth may limit scalability in real-time industrial deployment.

More recent approaches have incorporated attention mechanisms. BERT4NILM~\cite{bertnilm} leveraged bidirectional self-attention to capture global temporal dependencies, while PHCNN--GRU~\cite{PHCNN-GRU} combined parallel CNN feature extraction with recurrent temporal modeling for multi-task disaggregation. Energformer~\cite{b12} introduced a Transformer-based architecture with data-driven window selection to reduce manual tuning. Despite their modeling flexibility, attention-based and hybrid architectures remain computationally demanding and often require training separate models per appliance in industry.

Collectively, prior work reveals a persistent trade-off between temporal modeling capacity and computational efficiency. Lightweight convolutional models such as Seq2Point provide favorable inference latency but limited explicit receptive-field expansion, whereas recurrent, dilated, or attention-based architectures improve accuracy at significantly higher computational cost. Moreover, most studies focus on single-task disaggregation, leaving multi-task one-to-many industrial settings comparatively underexplored.

Building upon these observations and the comparative evaluation results (Table~\ref{tab:imdedl_comparison}), we introduce SEDR-Seq2P, a lightweight yet structurally enhanced Seq2Point architecture specifically designed for multi-task industrial NILM.

\section{Conclusion and Future Work}
In this paper, we propose a one-to-many model for multi-task energy disaggregation to enhance the effectiveness and practicality of NILM for large-scale industrial deployment. We benchmark Seq2Seq, Seq2SubSeq, Seq2Point, GRU, and WaveNet on IMDELD using complementary energy-estimation metrics and the Accuracy–Delay criterion. Seq2Point consistently outperforms Seq2Seq and Seq2SubSeq, indicating that point-wise prediction reduces redundancy and boundary effects. GRU and WaveNet achieve the highest disaggregation accuracy and event alignment, but at substantially higher training and inference cost, leading to weaker AccD trade-offs. In contrast, the proposed SEDR-Seq2P augments the Seq2Point backbone with dilated residual blocks and squeeze-and-excitation attention, providing consistent accuracy gains while preserving low inference latency and a favorable AccD balance for scalable industrial NILM deployment. This work supports Industry 4.0 by enabling edge-deployable, robust, data-driven monitoring of industrial equipment states and energy consumption from nonintrusive sensing.

As future work, we plan to explore adaptive sliding-window schemes that adjust context length to changing operating regimes under nonstationary conditions. We will also assess alternative NILM training objectives beyond MAE, with a particular focus on MR to better reflect event-level temporal alignment. In the longer term, we intend to improve generalization and deployability by incorporating additional electrical features, benchmarking robustness under increased measurement noise and load concurrency, evaluating transfer learning with cross-site validation, and studying lightweight compression strategies (e.g., pruning, quantization, and distillation) to enable low-latency edge deployment.


\begin{thebibliography}{00}


\bibitem{b8}
X.~Sun, J.~Hu, W.~Hu, D.~Cao, Z.~Chen, and F.~Blaabjerg, ``Non-intrusive load monitoring based on process-adaptive multi-target regression and transformer-enabled two-stream input network,'' \emph{Appl. Energy}, vol.~393, p.~126046, 2025, DOI: 10.1016/j.apenergy.2025.126046.

\bibitem{b9}
L.~Pereira and N.~Nunes, ``Performance evaluation in non-intrusive load monitoring: Datasets, metrics, and tools---A review,'' \emph{Wiley Interdiscip. Rev. Data Min. Knowl. Discov.}, vol.~8, no.~6, p.~e1265, 2018, DOI: 10.1002/widm.1265.

\bibitem{b10}
F.~Kalinke, P.~Bielski, S.~Singh, E.~Fouch{\'e}, and K.~B{\"o}hm, ``An evaluation of NILM approaches on industrial energy-consumption data,'' in \emph{Proc. ACM Int. Conf. Future Energy Syst. (e-Energy)}, 2021, pp.~239--243, DOI: 10.1145/3447555.3464863.

\bibitem{Yaniv}
A.~Yaniv and Y.~Beck, ``Advances in non-intrusive load monitoring for the industrial domain: Challenges, insights, and path forward,'' \emph{Renew. Sustain. Energy Rev.}, vol.~210, p.~115136, 2025, DOI: 10.1016/j.rser.2024.115136.

\bibitem{hart}
G.~W.~Hart, ``Nonintrusive appliance load monitoring,'' \emph{Proc. IEEE}, vol.~80, no.~12, pp.~1870--1891, Dec. 1992, DOI: 10.1109/5.192069.

\bibitem{Zhang}
C.~Zhang, M.~Zhong, Z.~Wang, N.~Goddard, and C.~Sutton, ``Sequence-to-point learning with neural networks for non-intrusive load monitoring,'' in \emph{Proc. AAAI Conf. Artif. Intell.}, 2018, DOI: 10.48550/arXiv.1612.09106.

\bibitem{Kelly}
J.~Kelly and W.~Knottenbelt, ``Neural NILM: Deep neural networks applied to energy disaggregation,'' in \emph{Proc. ACM Int. Conf. Embedded Syst. Energy-Efficient Built Environ. (BuildSys)}, 2015, pp.~55--64, DOI: 10.1145/2821650.2821672.

\bibitem{subseq}
Y.~Pan, K.~Liu, Z.~Shen, X.~Cai, and Z.~Jia, ``Sequence-to-subsequence learning with conditional GAN for power disaggregation,'' in \emph{Proc. IEEE Int. Conf. Acoust., Speech Signal Process. (ICASSP)}, 2020, pp.~3202--3206, DOI: 10.1109/ICASSP40776.2020.9053947.

\bibitem{ARR3}
S.~M.~Abdulrahman, P.~Brazdil, J.~N.~van~Rijn, and J.~Vanschoren, ``Speeding up algorithm selection using average ranking and active testing by introducing runtime,'' \emph{Mach. Learn.}, vol.~107, no.~1, pp.~79--108, 2018, DOI: 10.1007/s10994-017-5687-8.

\bibitem{ARR}
W.~Kong, Z.~Y.~Dong, B.~Wang, J.~Zhao, and J.~Huang, ``A practical solution for non-intrusive type II load monitoring based on deep learning and post-processing,'' \emph{IEEE Trans. Smart Grid}, vol.~11, no.~1, pp.~148--160, Jan. 2020, DOI: 10.1109/TSG.2019.2918330.

\bibitem{Jiang}
J.~Jiang, Q.~Kong, M.~D.~Plumbley, N.~Gilbert, M.~Hoogendoorn, and D.~M.~Roijers, ``Deep learning-based energy disaggregation and on/off detection of household appliances,'' \emph{ACM Trans. Knowl. Discov. Data}, vol.~15, no.~3, pp.~1--21, 2021, DOI: 10.1145/3441300.

\bibitem{Vavouris}
A.~Vavouris, L.~Stankovic, and V.~Stankovic, ``A non-intrusive load monitoring-enabled framework for load scheduling in the dairy industry,'' \emph{Appl. Energy}, vol.~398, p.~126456, 2025, DOI: 10.1016/j.apenergy.2025.126456.

\bibitem{residual}
K.~He, X.~Zhang, S.~Ren, and J.~Sun, ``Deep residual learning for image recognition,'' in \emph{Proc. IEEE Conf. Comput. Vis. Pattern Recognit. (CVPR)}, 2016, pp.~770--778, DOI: 10.1109/CVPR.2016.90.

\bibitem{dilated}
F.~Yu, V.~Koltun, and T.~Funkhouser, ``Dilated residual networks,'' in \emph{Proc. IEEE Conf. Comput. Vis. Pattern Recognit. (CVPR)}, 2017, pp.~472--480, DOI: 10.1109/CVPR.2017.75.

\bibitem{SE}
J.~Hu, L.~Shen, and G.~Sun, ``Squeeze-and-excitation networks,'' in \emph{Proc. IEEE Conf. Comput. Vis. Pattern Recognit. (CVPR)}, 2018, pp.~7132--7141, DOI: 10.1109/CVPR.2018.00745.

\bibitem{imdeld}
P.~B.~M.~Martins, V.~B.~Nascimento, A.~R.~de~Freitas, P.~B.~e~Silva, and R.~G.~D.~Pinto, ``Industrial Machines Dataset for Electrical Load Disaggregation (IMDELD),'' IEEE DataPort, Dec. 2018, DOI: 10.21227/cg5v-dk02.

\bibitem{zscore}
S.~G.~Patro and K.~K.~Sahu, ``Normalization: A preprocessing stage,'' \emph{arXiv preprint arXiv:1503.06462}, 2015, DOI: 10.48550/arXiv.1503.06462.

\bibitem{Correlation}
M.~Franzese and A.~Iuliano, ``Correlation analysis,'' in \emph{Encyclopedia of Bioinformatics and Computational Biology: ABC of Bioinformatics}, vol.~1. Elsevier, 2018, pp.~706--721, DOI: 10.1016/B978-0-12-809633-8.20358-0.

\bibitem{b12}
G.~F.~Angelis, C.~Timplalexis, A.~I.~Salamanis, S.~Krinidis, D.~Ioannidis, D.~Kehagias, and D.~Tzovaras, ``Energformer: A new transformer model for energy disaggregation,'' \emph{IEEE Trans. Consum. Electron.}, vol.~69, no.~3, pp.~308--320, Jan. 2023, DOI: 10.1109/TCE.2023.3237862.

\bibitem{Mayhorn}
E.~T.~Mayhorn, G.~P.~Sullivan, J.~M.~Petersen, R.~S.~Butner, and E.~M.~Johnson, ``Load disaggregation technologies: Real world and laboratory performance,'' Pacific Northwest National Laboratory, Richland, WA, USA, Tech. Rep. PNNL-SA-116560, 2016.

\bibitem{statisticaltests}
J.~Dem\v{s}ar, ``Statistical comparisons of classifiers over multiple data sets,'' \emph{J. Mach. Learn. Res.}, vol.~7, pp.~1--30, 2006.

\bibitem{b13}
P.~B.~Martins, J.~G.~Gomes, V.~B.~Nascimento, and A.~R.~de~Freitas, ``Application of a deep learning generative model to load disaggregation for industrial machinery power consumption monitoring,'' in \emph{Proc. IEEE Int. Conf. Commun., Control, Comput. Technol. Smart Grids (SmartGridComm)}, 2018, pp.~1--6, DOI: 10.1109/SmartGridComm.2018.8587415.


\bibitem{bertnilm}
Z.~Yue, C.~R.~Witzig, D.~Jorde, and H.~A.~Jacobsen, ``BERT4NILM: A bidirectional transformer model for non-intrusive load monitoring,'' in \emph{Proc. Int. Workshop Non-Intrusive Load Monitoring (NILM)}, 2020, pp.~89--93. DOI: 10.1145/3427771.3429390.


\bibitem{PHCNN-GRU}
J.~Ouzine, M.~Marzouq, S.~Dosse~Bennani, K.~Lahrech, and H.~El~Fadili, ``New parallel hybrid PHCNN--GRU deep learning model for multi-output NILM disaggregation,'' \emph{Energy Efficiency}, vol.~18, no.~3, pp.~1--27, 2025, DOI: 10.1007/s12053-025-10308-2.

\end{thebibliography}
\end{document}